\documentclass[sigconf]{acmart}
\usepackage[utf8]{inputenc}
\usepackage{graphicx}
\usepackage{subcaption}
\usepackage{adjustbox}
\usepackage{booktabs} 
\usepackage{tabularray}
\usepackage{makecell}
\usepackage{multirow}
\usepackage{multicol}
\usepackage{rotating}
\usepackage{enumitem}
\usepackage{tabularx}
\usepackage[normalem]{ulem}

\theoremstyle{plain}
\newtheorem{thm}{Theorem}
\theoremstyle{definition}
\newtheorem{defn}[thm]{Definition} 

\author{Jingwei Zuo}
\orcid{}
\affiliation{Technology Innovation Institute}
\email{Jingwei.Zuo@tii.ae}

\author{Wenbin Li}
\orcid{}
\affiliation{Technology Innovation Institute}
\email{Wenbin.Li@tii.ae}

\author{Michele Baldo}
\orcid{}
\affiliation{Technology Innovation Institute}
\email{Michele.Baldo@tii.ae}

\author{Hakim Hacid}
\orcid{123}
\affiliation{Technology Innovation Institute}
\email{Hakim.Hacid@tii.ae}

\copyrightyear{2023} 
\acmYear{2023} 
\setcopyright{acmlicensed}\acmConference[SIGSPATIAL '23]{The 31st ACM International Conference on Advances in Geographic Information Systems}{November 13--16, 2023}{Hamburg, Germany}
\acmBooktitle{The 31st ACM International Conference on Advances in Geographic Information Systems (SIGSPATIAL '23), November 13--16, 2023, Hamburg, Germany}
\acmPrice{15.00}
\acmDOI{10.1145/3589132.3625575}
\acmISBN{979-8-4007-0168-9/23/11}

\begin{document}
\pagestyle{empty}
\title{Unleashing Realistic Air Quality Forecasting: Introducing the Ready-to-Use \texttt{PurpleAirSF} Dataset}

\begin{abstract}
Air quality forecasting has garnered significant attention recently, with data-driven models taking center stage due to advancements in machine learning and deep learning models. However, researchers face challenges with complex data acquisition and the lack of open-sourced datasets, hindering efficient model validation.
This paper introduces \texttt{PurpleAirSF}, a comprehensive and easily accessible dataset collected from the PurpleAir network. With its high temporal resolution, various air quality measures, and diverse geographical coverage, this dataset serves as a useful tool for researchers aiming to develop novel forecasting models, study air pollution patterns, and investigate their impacts on health and the environment.
We present a detailed account of the data collection and processing methods employed to build \texttt{PurpleAirSF}. Furthermore, we conduct preliminary experiments using both classic and modern spatio-temporal forecasting models, thereby establishing a benchmark for future air quality forecasting tasks.
\end{abstract}

\vspace{-1em}


\vspace{-1em}


\maketitle

\section{Introduction}
Air quality is directly related to public health and environmental decision-making, and developing accurate and reliable air quality forecasting models is critical due to the high impact on living condition, ecosystem integrity, and climate change study. Modern air quality forecasting approaches, which leverage data-driven models~\cite{Han2022}, require large volumes of data. A reliable forecasting solution relies heavily on the availability of clean (i.e., low data noise), complete (i.e., minimum missing value), and contextually rich (i.e., integration of additional information such as meteorological data) datasets for training, validation and testing ~\cite{Mendez2023}. As such, the pursuit of accurate air quality forecasting is as much a data challenge as it is a methodological one, underlining the necessity for high-quality and ready-to-use datasets in this field.

However, the preparation of such datasets is resource consuming, and thus the availability is generally low. Some work with promising air quality results ~\cite{Liang2022} are realized upon proprietary datasets with limited accessibility. Such efforts, on one hand, prevent the reproducibility and restrains the applicability, and on the other hand present a hurdle for comparative studies.

Alternatively, open datasets~\cite{misc_beijing_multi-site_air-quality_data_501, misc_air_quality_360, zheng2013u} and APIs~\cite{OpenAQ2023, AQI2023, EPA2018} exist to foster the research progress of air quality forecasting, and allow researchers to validate models, refine methods, and generate findings that can be applied to real-world scenarios for air quality management. However, the open datasets fall short in the number of the included sensors to support the study large scale air quality evolution, or lack of the contextual data ~\cite{misc_beijing_multi-site_air-quality_data_501, misc_air_quality_360} to capture the hidden factors (e.g., meterological data and human activity data) influencing the air quality evolution. While open APIs from air quality monitoring platforms offer air quality data from a large number of available sensors cross country and continent, they still have several limitations such as no contextual data\cite{OpenAQ2023}, limited temporal sampling frequency\cite{EPA2018}, short historical period\cite{AQI2023}. In addition to these, the common barrier to use open APIs is the handling of the complex APIs for data retrieval, and the varying data quality (e.g., missing values or abnormal values) across sensors due to malfunction or disconnection. As a result, the limitations of the datasets and APIs narrow the scope of the research opportunities that can be addressed and constrains the potential to fully explore the air quality forecasting solutions.

Against the drawbacks, we present a high-quality, ready-to-use dataset\footnote{https://github.com/JingweiZuo/PurpleAirSF} for air quality forecasting, sourced from the city of San Francisco. The dataset was initially obtained with PurpleAir API~\cite{PurpleAir2023} and subsequently processed. 
It combines the benefits of open datasets and APIs, resulting in several key features:
\begin{itemize}[leftmargin=1em]
    \item \textbf{Fine-grained spatial resolution}. The provided dataset is built on a substantial number of air quality sensors (up to 316 sensor stations in San Francisco), leading to a dense spatial distribution.
    \item \textbf{Multiple high-sampling rates}. It offers three high sampling frequencies (10-minute, 1-hour, and 6-hour intervals) spanning over 1.5 years, with up to 3\% missing values.
    \item \textbf{Rich contextural information}. Complemented with meteorological data and light measurements collected at the same frequencies, the dataset provides a comprehensive insight into the various factors influencing air quality.
\end{itemize}


The rest of the paper is organized as follows: Section 2 reviews relevant open datasets and open APIs. Section 3 provides a detailed description of our dataset, while Section 4 presents preliminary results on spatio-temporal forecasting models on our dataset. Section 5 outlines potential research problems addressed by our dataset. 
 






\section{Related work}
Open datasets and APIs offer the valuable access to a wide variety of air quality metrics across different spatial and temporal scales. 

\subsection{Open Datasets}
Table~\ref{tab:Open datasets} shows an overview of popular open datasets on air quality data. The Italian single-site air quality dataset~\cite{misc_air_quality_360} combines five oxide chemical sensors with pollutants values (e.g., $NO_x$), representing the longest open recordings of on field deployed air quality chemical data. Alternatively, the Beijing multi-site air quality dataset~\cite{misc_beijing_multi-site_air-quality_data_501} collects six Air Quality Index (AQI) measures and six meteorological data from 12 monitoring sites. Similarly, U-Air~\cite{zheng2013u} is a collection of datasets, with AQI and meteorological data sourced from Beijing, Guangzhou, Tianjin, Shenzhen. 

\begin{table}[!htbp]
\centering
\vspace{-1em}
\caption{Overview of popular open Air Quality datasets}
\vspace{-1em}
\label{tab:Open datasets}
\scalebox{0.75}{
\begin{tabular}{llllll}
\toprule
Open dataset & \#Stations & Measures                     & Frequency & Archive   \\
\midrule
Italien AQ   & 1          & Chemical measures, $CO_x$    & Hourly    & 2004-2005 \\
Beijing AQ   & 12         & AQI, Meteorology ($aligned$)   & Hourly    & 2013-2017 \\
U-Air        & 11-42         & AQI, Meteorology ($unaligned$) & Hourly    & 2014-2015 \\
\bottomrule
\end{tabular}
}
\vspace{-1em}
\end{table}
While current open datasets are useful, they exhibit limitations. These encompass sparse sensor coverage, hindering large-scale studies, a lack of contextual data, and uniform sampling frequency. U-Air~\cite{zheng2013u}, a popular choice for air quality forecasting, necessitates careful data alignment between AQI and meteorological measures. However, this can introduce inaccuracies due to sparse sensor coverage, potentially undermining model performance. Consequently, while such datasets are a good starting point, their inherent limitations constrain their capacity to address extended research queries.


\subsection{Open APIs}
Open APIs provide an avenue for data retrieval from air quality monitoring platforms and enable access to data from numerous sensors across various locations, thus facilitating large scale studies. 


Table \ref{tab:Open APIs} compares the popular Open APIs in terms of spatial \textit{coverage}, \textit{context} (i.e., integration of contextual data), sampling \textit{frequency}, \textit{archive} (i.e., the historical data retrieval period), and data \textit{quality} (i.e., presence and percentage of missing values and abnormal values). As APIs integrate sensors from various sources, the data quality largely varies among the integrated sensors (e.g, up to 50\% missing values for specific sensor), requiring a careful filtering and selection when using the APIs. 
In this work, we choose to use the PurpleAir API~\cite{PurpleAir2023} for the initial preparation of our dataset owing to its finest data sampling frequency (i.e., every minute) for customized requirements, long historical availability, and rich context integration (i.e., meteorology and light measures). 

\begin{table}[!htbp]
\centering
\caption{Overview of popular open APIs for Air Quality data}
\vspace{-1em}
\label{tab:Open APIs}
\scalebox{0.75}{
\begin{tabular}{llllll}
\toprule
Open APIs   & Coverage & Context & Frequency & Archive & Quality \\
\midrule
OpenAQ~\cite{OpenAQ2023}      & Global     & No      &    Hour    &  Up to 2015   &    Varying          \\
AirNow~\cite{EPA2018}      & USA        &  No     &    Hour    &  1 year       &    Varying            \\
WAQI~\cite{AQI2023}        & Global     &  Yes    &    Hour    & Up to 2015   &    Varying          \\
PurpleAir~\cite{PurpleAir2023}   & Global     &  Yes    &    Minute  &  Up to 2015   &    Varying    \\      
\bottomrule
\end{tabular}
}
\vspace{-1em}
\end{table}

However, open APIs pose their own challenges. Firstly, dealing with complex APIs for data retrieval can be a significant technical barrier for researchers. Secondly, the varying data quality among the integrated sensors necessitates careful filtering and selection, leading to time-consuming preprocessing stages. These factors can make it difficult to use open APIs directly.

Given these limitations of existing open datasets and APIs, there is a clear need for a data solution that amalgamates the advantages of both. 
Such a solution would incorporate numerous sensors at fine-grained temporal frequency, provide rich contextual data to capture hidden factors, allow for long historical records, assure high data quality, and most importantly, be ready-to-use with minimal preprocessing efforts.

\section{PurpleAirSF - A ready-to-use dataset on PurpleAir}

In this section, we present \texttt{PurpleAirSF}, a ready-to-use dataset built on top of the PurpleAir API, with careful preprocessing and transformations. 
We show how the data was collected, processed and transformed into the desired format to feed forecasting models.

\subsection{Raw Data Acquisition}

Owning to the extensive information and broad coverage provided by PurpleAir, we utilized the PurpleAir API to collect raw data.
PurpleAir provides open access to air quality data amassed by its monitoring network. This data encompasses measurements of particulate matter (PM2.5 and PM10), temperature, humidity, among others. To ensure the reliability of their data, PurpleAir has implemented quality control measures.

Air pollution strongly correlates with weather and meteorological conditions. Hence, both the air quality index (AQI) data and meteorological information are considered in our data collection process. 
By default, PurpleAir API provides various options of data granularity or sampling rates: one day, six hours, one hour, and 10 minutes. Higher granularity data is generally an aggregation of data with lower granularity.
Compared to other spatio-temporal data (e.g., traffic), air quality data exhibit unique features:
\begin{itemize}[leftmargin=1em]
\item Multiple external factors should be considered for accurate air quality predictions, including weather conditions, urban planning, traffic, surrounding human activities, and more.
\item Unlike the discrete nature between traffic nodes, air quality data is continuous in spatial space~\cite{lin2022conditional}, making air quality forecasting a more challenging research problem than traffic forecasting.
\end{itemize}

Therefore, we consider both meteorological data and air quality data during the data collection process. Further, considering the continuous spatial features in air quality data, we chose to rely on data with high spatial resolution.
Specifically, we focused on air quality monitoring and forecasting at the city scale, choosing to collect data from the center of \textbf{San Francisco}. 
This area, covering approximately 10 x 10 square kilometers, brings together a total of 638 sensors for monitoring and analysis, providing a rich and high-resolution dataset for air quality forecasting research.

Given the periodic features exhibited by air quality data, similar to other spatio-temporal data like traffic~\cite{zuo2023graph}, it's beneficial to have a long-term dataset. For this reason, we retrieved all available data for the past five years (\textit{2018-05-16} to \textit{2023-05-15}). This serves as the raw data for our subsequent processing.
\subsection{Data Preprocessing}
We utilize the historical API provided by PurpleAir, which encompasses diverse sensor measurements related to air quality. Out of these, we have selected 19 measurements that directly indicate the quality of air and external factors affecting it. As delineated in Table~\ref{tab:aq_measures}, these selected measurements fall into five categories: meteorological (MET.) measures, particle counts, particulate matters, light measures, and positions. This differs from previously available open-source datasets (e.g., Beijing dataset~\cite{misc_beijing_multi-site_air-quality_data_501}, U-Air~\cite{zheng2013u}), which consider meteorological information at a district or city level. Our collected data facilitates spatial alignment between meteorological and air quality measures at the sensor station level, sparing users the extensive work of data alignment. 

\begin{table}[!htbp]
\centering
\caption{Selected Air Quality measures from PurpleAir APIs}
\vspace{-1em}
\scalebox{0.8}{
\begin{tabular}{m{1.5cm}m{2.5cm}m{5.7cm}}
\toprule
\multicolumn{1}{c}{Categories} & \multicolumn{1}{c}{Measures} & \multicolumn{1}{c}{Descriptions} \\
\midrule
\multirow{3}{*}{\makecell{MET.\\measures}} & - humidity & - Relative humidity inside sensor housing (\%) \\
& - temperature & - Temperature inside the sensor housing (F) \\
& - pressure & - Current pressure in Millibars \\
\midrule
\makecell{Particle\\counts} & - \{size\}\_um\_count & - Count concentration (particles/100ml) of all particles greater than size = 0.3, 0.5, 1.0, 2.5, 5.0, 10.0 µm diameter. \\
\midrule
\makecell{Particulate\\matters} & \makecell[l]{- pm1.0\_\{cf\_1,atm\} \\ - pm2.5\_\{cf\_1,alt,atm\} \\ - pm10.0\_\{cf\_1,atm\} } & - Various variants (e.g., cf=1, atm, alt) for related Fine Particles measures, which are derived from the particle counts \\
\midrule
\multirow{3}{*}{\makecell{Light\\measures}} & - scattering\_coeff. & - Fine Particulate Light Scattering \\
& - deciviews & - A haze index related to light scattering and extinction that is approximately linearly related to human perception of the haze \\
& - visual\_range & - Referred to Visibility, the visual range is the distance from the observer that a large dark object just disappears from view. \\
\midrule
Positions & \makecell[l]{- longtitude\\- latitude} & - The GPS locations of the sensors. \\
\bottomrule
\end{tabular}
}
\vspace{-1em}
\label{tab:aq_measures}
\end{table}

\begin{defn}(Preprocessed data format).
The collected data is represented by a \textit{DataFrame} (i.e., tensor) with shape $(N,L,F)$, where $N$ is the number of sensor stations, $L$ is the sequence length, $F$ is the number of air quality measures. 
\end{defn}

Basically, the air quality data in each sensor station can be regarded as a Multivariate Time Series (MTS)~\cite{zuo2021smate}. 
The collected raw data, however, presents multiple data quality issues, e.g., outliers and missing values, which can arise due to communication errors or sensor failures, and they directly impact forecasting models.

\begin{figure*}[htbp]
    \begin{adjustbox}{width=1.05\textwidth,center}
    \centering
    \begin{subfigure}{.22\linewidth}
        \includegraphics[width=\linewidth]{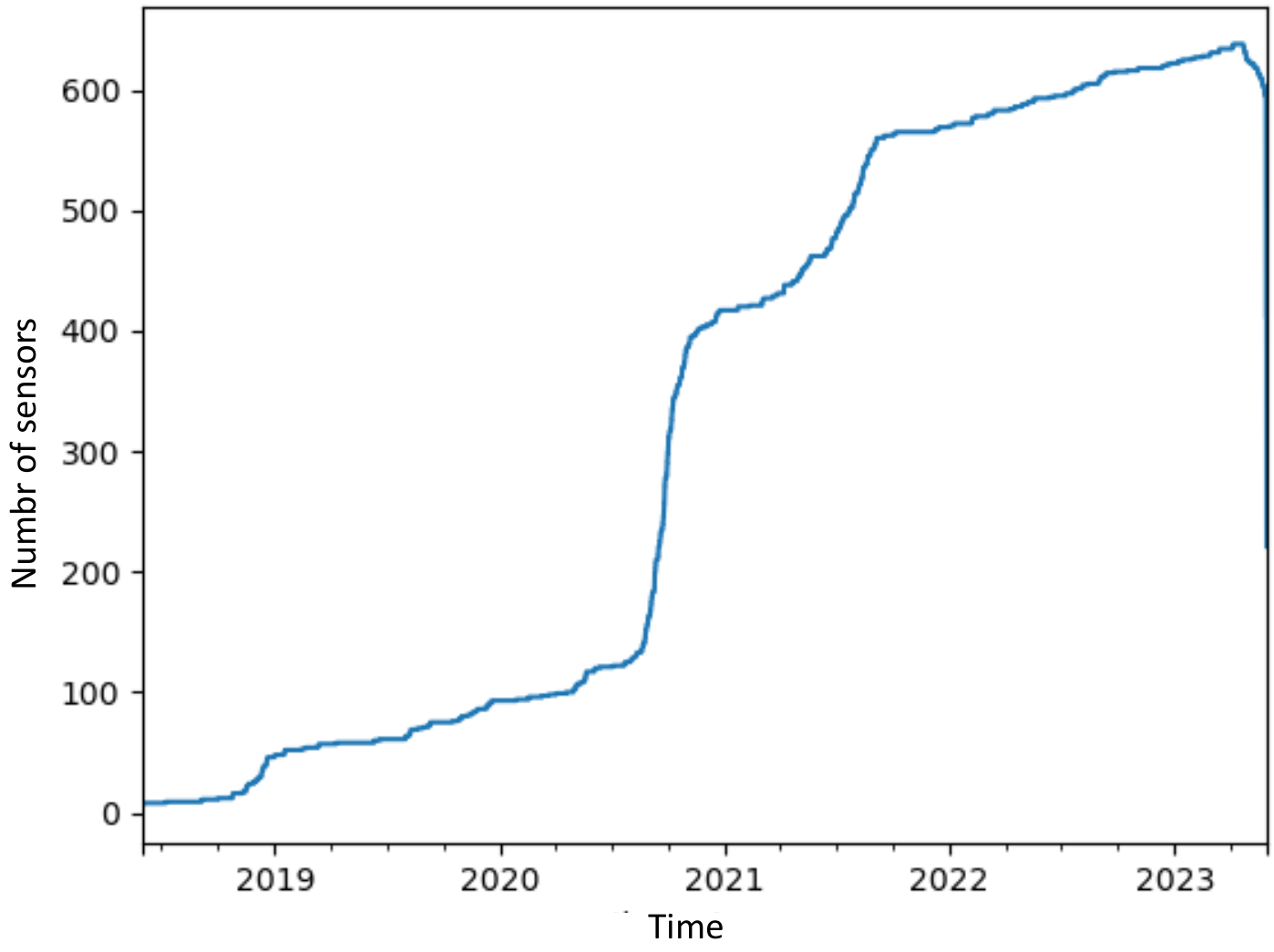}
        \caption{\textit{Nbr. of stations} over \textit{time}}
        \label{fig:missing_data_overview:a}
    \end{subfigure}
    \hfill
    \begin{subfigure}{.22\linewidth}
        \includegraphics[width=\linewidth]{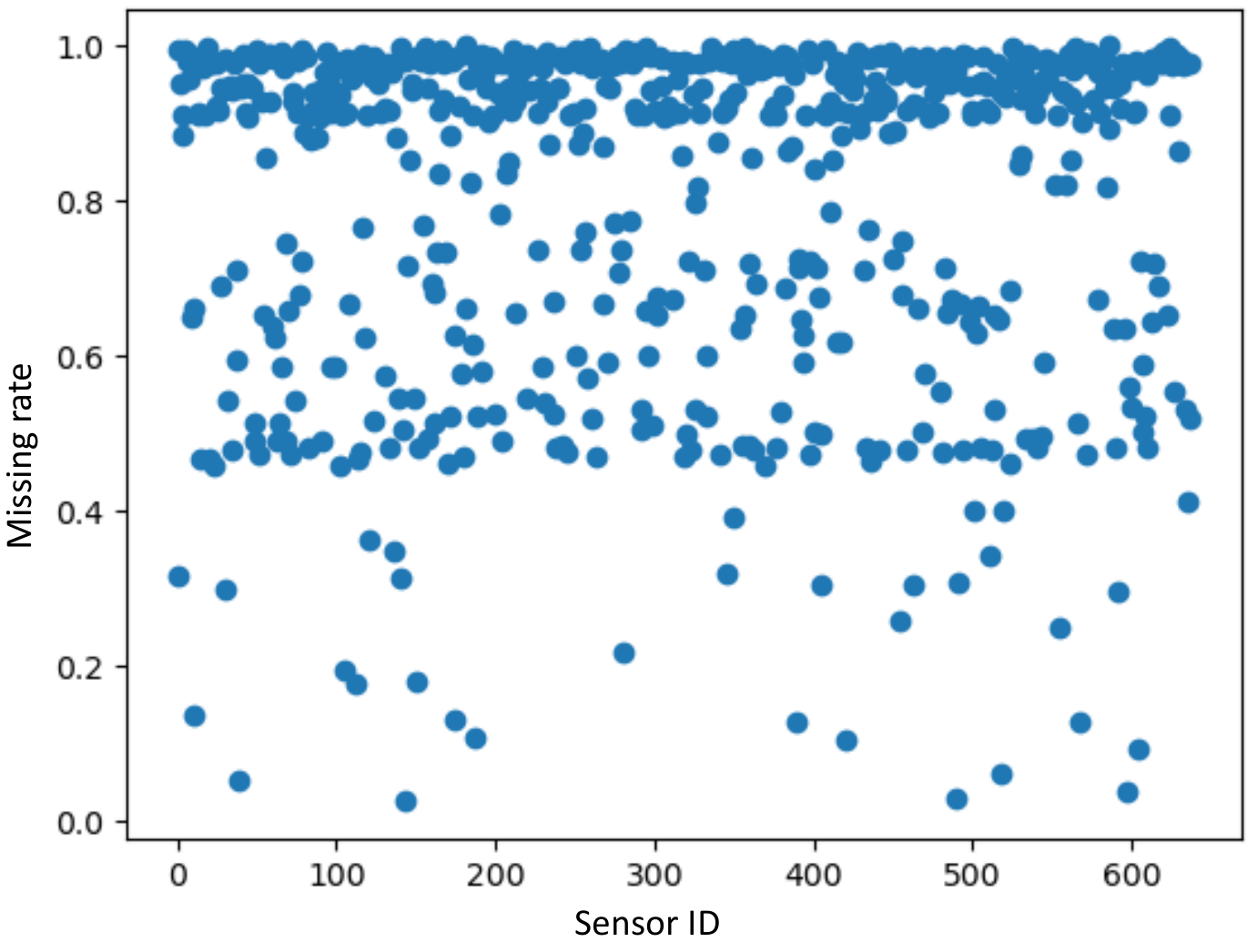}
        \caption{\textit{Missing rate} over \textit{stations}}
        \label{fig:missing_data_overview:b}
    \end{subfigure}
    \hfill
    \begin{subfigure}{.19\linewidth}
        \includegraphics[width=\linewidth]{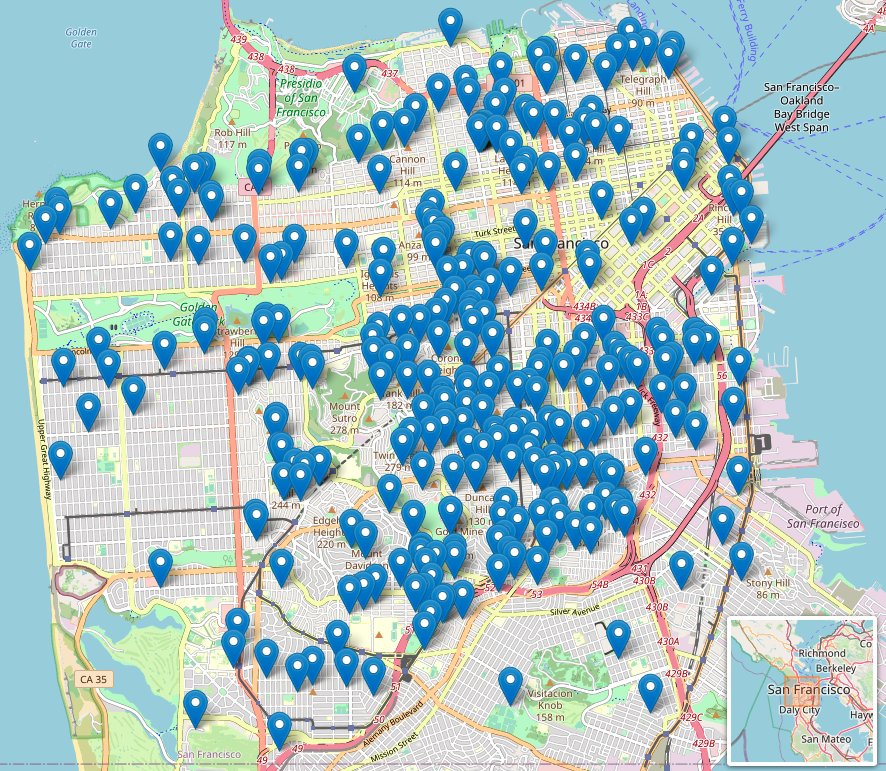}
        \caption{GPS -316 stations (10M)}
        \label{fig:gps_10M_316}
    \end{subfigure}
    \hfill
    \begin{subfigure}{.19\linewidth}
        \includegraphics[width=\linewidth]{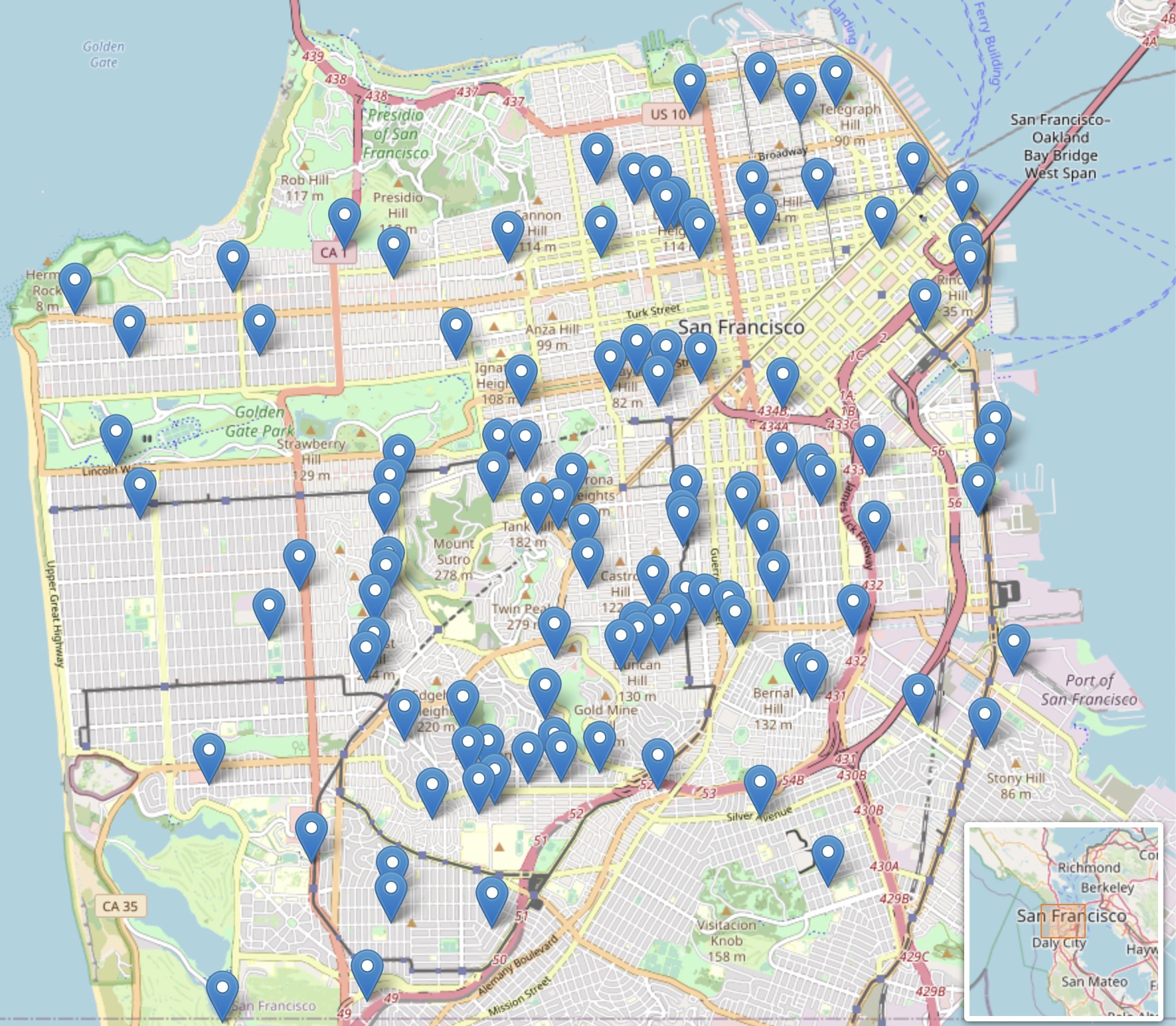}
        \caption{GPS -112 stations (1H)}
        \label{fig:gps_1H_112}
    \end{subfigure}
    \hfill
    \begin{subfigure}{.19\linewidth}
        \includegraphics[width=\linewidth]{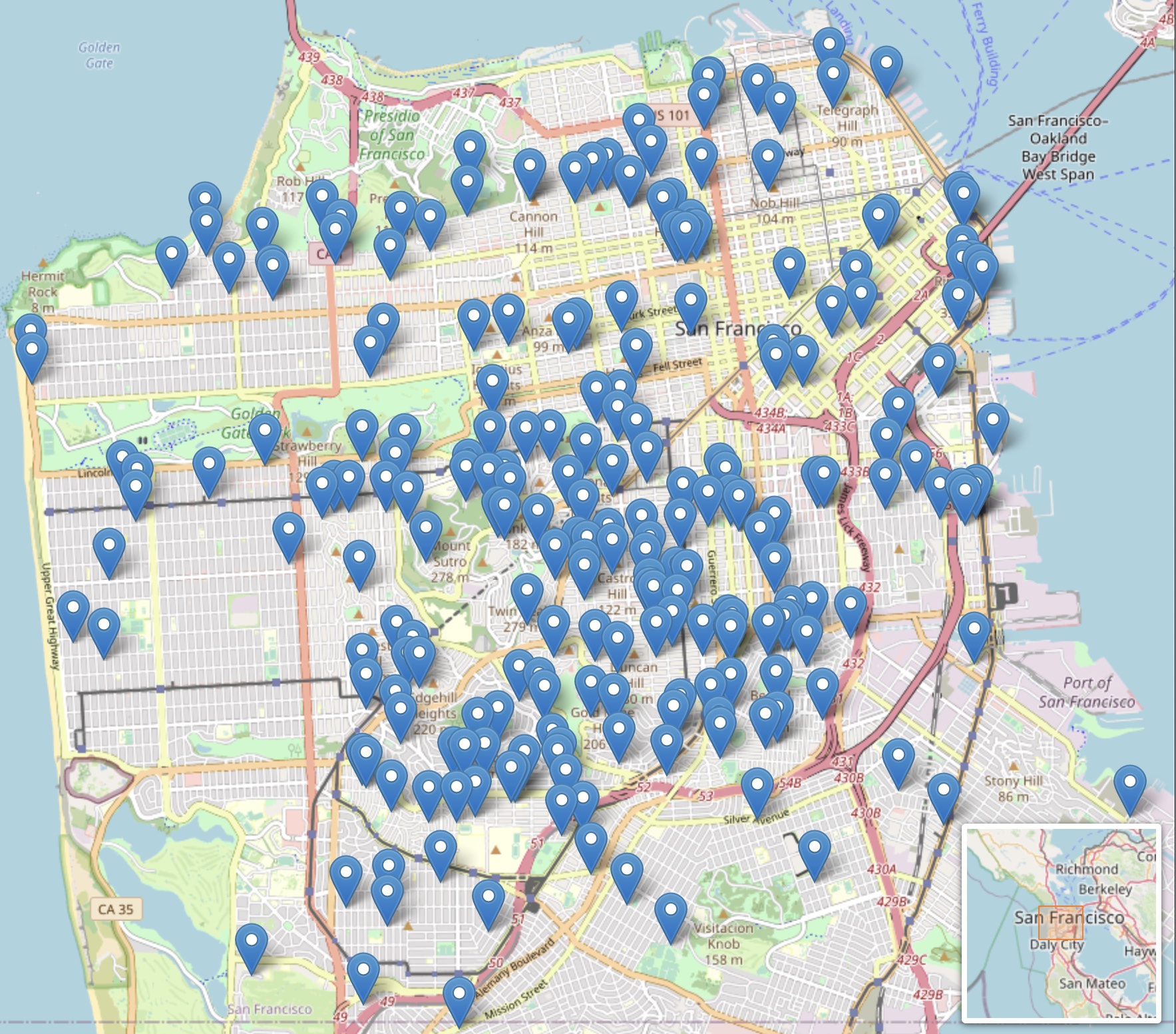}
        \caption{GPS -232 stations (6H)}
        \label{fig:gps_6H_232}
    \end{subfigure}
    \end{adjustbox}
    \vspace{-2em}
    \caption{\centering (a-b) Missing value overview for the raw data, collected from 638 stations between 16/05/2018 and 15/05/2023; (c-e) Sensor station distribution for preprocessed data (San Francisco central city)}
    \label{fig:2}
\vspace{-1em}
\end{figure*}

There are numerous methods to handle missing values during preprocessing, ranging from statistic imputations (e.g., mean, median)~\cite{zuo2023graph} to recent generative models~\cite{yoon2019time}. They can be applied to manage both short-range and long-range missing values in time series.
However, as shown in Figure~\ref{fig:missing_data_overview:a}, in the practical realm, we observe that PurpleAir expands gradually its sensor stations. This incremental expansion leads to a significant number of missing values over recently installed sensors (Figure~\ref{fig:missing_data_overview:b}).
To this end, we carefully observe the missing value distributions among both temporal and spatial axis, to filter out the most valuable time periods and sensor nodes. This careful filtration helps minimize the introduction of artificial data into the raw dataset, thereby minimizing potential interference with the original data's inherent information.

Concretely, we apply threshold-based filtering for the spatio-temporal data: i) for the temporal axis, as shown in Figure~\ref{fig:missing_data_overview:a}, we use only the recent one and half year's records, i.e., 2021-10-01 to 2023-05-15; ii) for the spatial axis, we disregard any sensors with a missing rate above a specified threshold (e.g., 2\%) within a given granularity. This is based on the recently recorded data. 

\vspace{-1em}
\begin{table}[!htbp]
\centering
\caption{Summary statistics of PurpleAirSF-10M/1H/6H}
\vspace{-1em}
\label{AQIDatasets}
\scalebox{0.75}{
\begin{tabular}{ccccccc}
\toprule
\textbf{Data} & \textbf{\#Stations} & \textbf{\#Features} & \textbf{Sampling} & \textbf{Observations} & \textbf{Missing} \\
\midrule

PurpleAirSF-10M & 316           & 19         & 10 mins  &  491 277 300            & 3.080\%        \\
PurpleAirSF-1H    & 112        & 19         & 1 hour   & 29 011 024   & 1.566\% \\
PurpleAirSF-6H      & 232        & 19         & 6 hours  & 10 054 648   & 1.231\% \\
\bottomrule
\end{tabular}
}
\label{tab:data_statistics}
\vspace{-1em}
\end{table}

To identify and filter outliers, we employ Z-score measures~\cite{blazquez2021review}, which quantifies a data point's relationship to the mean of a group of values in terms of standard deviations. We consider data points with an absolute Z-score exceeding 3 as outliers and replace them with the maximum normal value. 
Table~\ref{tab:data_statistics} provides an overview for the processed multi-granular datasets (10-min, one \& six-hour), while Figure~\ref{fig:gps_10M_316},\ref{fig:gps_1H_112},\ref{fig:gps_6H_232} show the sensor distributions over the city.


\section{Benchmark Results}
To gain deeper insights into the provided datasets, we employ several machine learning models for an initial performance comparison. The datasets are divided into training, validation, and test sets in a \textit{7:1:2} ratio. We use the data from the last 12 time steps to predict the AQI (i.e., PM2.5) for the subsequent 12 steps. The forecasting accuracy of all models tested is evaluated using three metrics~\cite{zuo2023graph}: mean absolute error (MAE), root-mean-square error (RMSE), and mean absolute percentage error (MAPE). For this test, we selected two widely-used models: \textbf{LSTM} (Long Short-Term Memory) and \textbf{GWN} (Graph WaveNet)~\cite{wu2019graph}. By taking into account (or not) the meteorological measures, we gain a better understanding of the data's characteristics.

Table~\ref{tab:benchmark_results} presents the benchmark results across different forecasting horizons (i.e., steps), which show several insights: 
\begin{itemize}[leftmargin=1em]
    \item As the forecasting horizon extends, the model performance tends to degrade due to limited knowledge propagation over time; 
    \item The intricate spatial relationships among sensor stations are pivotal to air quality forecasting. By incorporating these relationships (e.g., through Graph models), the GWN has significantly enhanced the forecasting performance; 
    \item The performance of most models generally improves with fine-granularity data. However, when the time granularity decreases to 10 minutes, the performance of LSTM significantly deteriorates, while GWN shows improved results. This can be attributed to the fact that making use of the spatial information offsets the impact of fine-granular temporal dynamics. This phenomenon highlights the importance of incorporating spatial information while dealing with high-frequency, fine-granular air quality data.
    \item Incorporating meteorological measures, has enabled the GWN to enhance its forecasting performance. However, surprisingly, the performance of LSTM deteriorates upon this integration. This is due to the fact that the incorporation increases the model's complexity, which in turn leads to overfitting on the training data. A model designed more effectively could take advantage of the rich meteorological data to boost its forecasting performance. 
\end{itemize}


\begin{table}[t]
\centering
\caption{Performance comparison over different horizons.}
\vspace{-1em}
\label{tab:benchmark_results}
\scalebox{0.65}{
\begin{tabular}{clp{0.7cm}p{0.7cm}p{0.8cm}p{0.7cm}p{0.7cm}p{0.8cm}p{0.7cm}p{0.7cm}p{0.8cm}}
\toprule
                &       & \multicolumn{3}{c}{Horizon=3}                                   & \multicolumn{3}{c}{Horizon=6}                                    & \multicolumn{3}{c}{Horizon=12}                                  \\
                
          & Models                & MAE                 & RMSE          & MAPE(\%)        & MAE                 & RMSE          & MAPE(\%)       & MAE                 & RMSE          & MAPE(\%)         \\
\midrule
\multirow{3}{*}{\rotatebox{90}{P.SF-10M}}    

  & LSTM (No-Meteo) & 6.86          & 22.43          & 8.23          & 7.21          & 13.23          & 8.76           & 7.54          & 14.22          & 9.22           \\
 & GWN (No-Meteo)  & \uline{1.67}  & \uline{8.84}   & \uline{1.95}  & \uline{2.14}  & \uline{8.47}   & \uline{1.86}   & \uline{2.22}  & \uline{8.98}   & \uline{1.73}   \\
 & LSTM            & 7.84          & 26.03          & 10.56         & 7.55          & 17.86          & 9.01           & 10.38         & 16.38          & 9.35           \\
 & GWN             & \textbf{1.66} & \textbf{8.67}  & \textbf{1.74} & \textbf{2.03} & \textbf{8.46}  & \textbf{1.65}  & \textbf{2.13} & \textbf{8.50}  & \textbf{1.39}  \\

\midrule
\multirow{3}{*}{\rotatebox{90}{P.SF-1H}} & LSTM (No-Meteo) & 3.09          & 12.86          & 3.81          & 3.77          & 11.57          & 6.77           & 4.33          & 12.65          & \uline{10.36}   \\
 & GWN (No-Meteo)  & \uline{2.83}  & \uline{9.56}   & 4.36          & \uline{3.67}  & \uline{11.11}  & 10.30          & \uline{4.23}  & \uline{11.98}  & 10.49          \\
 & LSTM            & 3.38          & 14.86          & \uline{3.95}  & 3.81          & 14.91          & \uline{6.30}   & 4.37          & 15.91          & 8.79           \\
 & GWN             & \textbf{2.76} & \textbf{9.55}  & \textbf{4.05} & \textbf{3.57} & \textbf{11.10} & \textbf{7.62}  & \textbf{4.18} & \textbf{11.96} & \textbf{9.59}  \\

\midrule
 \multirow{3}{*}{\rotatebox{90}{P.SF-6H}}      & LSTM (No-Meteo) & 8.00          & 33.27          & 11.96         & 9.01          & 35.32          & 17.30          & 7.95          & 32.88          & \uline{10.96}  \\
 & GWN (No-Meteo)  & \uline{4.34}  & \uline{12.77}  & \uline{11.57} & \uline{5.07}  & \uline{14.93}  & 17.23          & \uline{4.71}  & \uline{15.03}  & 14.47          \\
 & LSTM            & 9.36          & 36.80          & 16.53         & 8.72          & 33.12          & \uline{12.70}  & 8.96          & 35.96          & 19.10          \\
 & GWN             & \textbf{4.31} & \textbf{11.61} & \textbf{9.73} & \textbf{5.04} & \textbf{14.92} & \textbf{12.63} & \textbf{4.64} & \textbf{14.91} & \textbf{13.39}         \\

\bottomrule
\end{tabular}
}
\vspace{-2em}
\end{table}
\section{Research Opportunities}
From the benchmark results, it's clear that we can have multiple research opportunities (RO) on the provided datasets:
\begin{itemize}
\item[\textbf{RO1}] \textit{Meteorological Data Exploration}: Meteorological data and air quality measures should have distinct contribution weights in air quality forecasting. An effective model architecture should strike a balance between model complexity and performance when making use of meteorological data.
\item[\textbf{RO2}] \textit{Spatial Relationship Modeling}: Air quality inherently exhibits continuous spatial relationships~\cite{lin2022conditional}, making it more complex than other spatio-temporal data such as city traffics.
\item[\textbf{RO3}] \textit{Long-term Forecasting}: Given the lengthy time of the collected data, using more historical data to recognize longer historical patterns could further optimize the forecaster.
\end{itemize}

Numerous research opportunities can be explored using the provided multi-granular datasets. Depending on specific context requirements (e.g., forecasting in complex context with missing values~\cite{zuo2023graph}), these datasets can be adjusted accordingly.



\section{Conclusion}
In this paper, we introduced PurpleAirSF, a ready-to-use dataset tailored for realistic air quality forecasting. With three distinct granularity levels - 10 minutes, one hour, and six hours - the datasets include meteorological data alongside comprehensive air quality measures. 
Distinct from existing open-source datasets, PurpleAirSF provides high spatial resolution, low rates of missing data, and fine granularity, facilitating a wide array of research problems.

\bibliographystyle{ACM-Reference-Format}
\bibliography{sample-base}


\begin{thebibliography}{16}


\ifx \showCODEN    \undefined \def \showCODEN     #1{\unskip}     \fi
\ifx \showDOI      \undefined \def \showDOI       #1{#1}\fi
\ifx \showISBNx    \undefined \def \showISBNx     #1{\unskip}     \fi
\ifx \showISBNxiii \undefined \def \showISBNxiii  #1{\unskip}     \fi
\ifx \showISSN     \undefined \def \showISSN      #1{\unskip}     \fi
\ifx \showLCCN     \undefined \def \showLCCN      #1{\unskip}     \fi
\ifx \shownote     \undefined \def \shownote      #1{#1}          \fi
\ifx \showarticletitle \undefined \def \showarticletitle #1{#1}   \fi
\ifx \showURL      \undefined \def \showURL       {\relax}        \fi
\providecommand\bibfield[2]{#2}
\providecommand\bibinfo[2]{#2}
\providecommand\natexlab[1]{#1}
\providecommand\showeprint[2][]{arXiv:#2}

\bibitem[\protect\citeauthoryear{{AQI {\&} Health | AirNow.gov}}{{AQI {\&}
  Health | AirNow.gov}}{2023}]%
        {AQI2023}
\bibfield{author}{\bibinfo{person}{{AQI {\&} Health | AirNow.gov}}.}
  \bibinfo{year}{2023}\natexlab{}.
\newblock
\newblock
\urldef\tempurl%
\url{https://www.airnow.gov/aqi-and-health/}
\showURL{%
\tempurl}


\bibitem[\protect\citeauthoryear{Bl{\'a}zquez-Garc{\'\i}a, Conde, Mori, and
  Lozano}{Bl{\'a}zquez-Garc{\'\i}a et~al\mbox{.}}{2021}]%
        {blazquez2021review}
\bibfield{author}{\bibinfo{person}{Ane Bl{\'a}zquez-Garc{\'\i}a},
  \bibinfo{person}{Angel Conde}, \bibinfo{person}{Usue Mori}, {and}
  \bibinfo{person}{Jose~A Lozano}.} \bibinfo{year}{2021}\natexlab{}.
\newblock \showarticletitle{A review on outlier/anomaly detection in time
  series data}.
\newblock \bibinfo{journal}{\emph{ACM Computing Surveys (CSUR)}}
  \bibinfo{volume}{54}, \bibinfo{number}{3} (\bibinfo{year}{2021}),
  \bibinfo{pages}{1--33}.
\newblock


\bibitem[\protect\citeauthoryear{Chen}{Chen}{2019}]%
        {misc_beijing_multi-site_air-quality_data_501}
\bibfield{author}{\bibinfo{person}{Song Chen}.}
  \bibinfo{year}{2019}\natexlab{}.
\newblock \bibinfo{title}{{Beijing Multi-Site Air-Quality Data}}.
\newblock \bibinfo{howpublished}{UCI Machine Learning Repository}.
\newblock
\newblock
\shownote{{DOI}: https://doi.org/10.24432/C5RK5G.}


\bibitem[\protect\citeauthoryear{EPA}{EPA}{2018}]%
        {EPA2018}
\bibfield{author}{\bibinfo{person}{EPA}.} \bibinfo{year}{2018}\natexlab{}.
\newblock \bibinfo{title}{{World's Air Pollution: Real-time Air Quality
  Index}}.
\newblock
\newblock
\urldef\tempurl%
\url{https://waqi.info/{\#}/c/6.769/8.931/2.1z{\%}0Ahttps://waqi.info/}
\showURL{%
\tempurl}


\bibitem[\protect\citeauthoryear{Han, Liu, Xiong, and Yang}{Han
  et~al\mbox{.}}{2022}]%
        {Han2022}
\bibfield{author}{\bibinfo{person}{Jindong Han}, \bibinfo{person}{Hao Liu},
  \bibinfo{person}{Haoyi Xiong}, {and} \bibinfo{person}{Jing Yang}.}
  \bibinfo{year}{2022}\natexlab{}.
\newblock \showarticletitle{{Semi-Supervised Air Quality Forecasting via
  Self-Supervised Hierarchical Graph Neural Network}}.
\newblock \bibinfo{journal}{\emph{IEEE Transactions on Knowledge and Data
  Engineering}} (\bibinfo{date}{may} \bibinfo{year}{2022}).
\newblock


\bibitem[\protect\citeauthoryear{Liang, Xia, Ke, Wang, Wen, Zhang, Zheng, and
  Zimmermann}{Liang et~al\mbox{.}}{2022}]%
        {Liang2022}
\bibfield{author}{\bibinfo{person}{Yuxuan Liang}, \bibinfo{person}{Yutong Xia},
  \bibinfo{person}{Songyu Ke}, \bibinfo{person}{Yiwei Wang},
  \bibinfo{person}{Qingsong Wen}, \bibinfo{person}{Junbo Zhang},
  \bibinfo{person}{Yu Zheng}, {and} \bibinfo{person}{Roger Zimmermann}.}
  \bibinfo{year}{2022}\natexlab{}.
\newblock \showarticletitle{{AirFormer: Predicting Nationwide Air Quality in
  China with Transformers}}.
\newblock  (\bibinfo{year}{2022}).
\newblock
\showeprint[arxiv]{2211.15979}


\bibitem[\protect\citeauthoryear{Lin, Gao, Xu, Wu, Li, and Li}{Lin
  et~al\mbox{.}}{2022}]%
        {lin2022conditional}
\bibfield{author}{\bibinfo{person}{Haitao Lin}, \bibinfo{person}{Zhangyang
  Gao}, \bibinfo{person}{Yongjie Xu}, \bibinfo{person}{Lirong Wu},
  \bibinfo{person}{Ling Li}, {and} \bibinfo{person}{Stan~Z Li}.}
  \bibinfo{year}{2022}\natexlab{}.
\newblock \showarticletitle{Conditional local convolution for spatio-temporal
  meteorological forecasting}. In \bibinfo{booktitle}{\emph{Proceedings of the
  AAAI Conference on Artificial Intelligence}}, Vol.~\bibinfo{volume}{36}.
  \bibinfo{pages}{7470--7478}.
\newblock


\bibitem[\protect\citeauthoryear{M{\'{e}}ndez, Merayo, and
  N{\'{u}}{\~{n}}ez}{M{\'{e}}ndez et~al\mbox{.}}{2023}]%
        {Mendez2023}
\bibfield{author}{\bibinfo{person}{Manuel M{\'{e}}ndez},
  \bibinfo{person}{Mercedes~G. Merayo}, {and} \bibinfo{person}{Manuel
  N{\'{u}}{\~{n}}ez}.} \bibinfo{year}{2023}\natexlab{}.
\newblock \showarticletitle{{Machine learning algorithms to forecast air
  quality: a survey}}.
\newblock \bibinfo{journal}{\emph{Artif. Intell. Rev.}} (\bibinfo{date}{feb}
  \bibinfo{year}{2023}), \bibinfo{pages}{1--36}.
\newblock


\bibitem[\protect\citeauthoryear{{OpenAQ}}{{OpenAQ}}{2023}]%
        {OpenAQ2023}
\bibfield{author}{\bibinfo{person}{{OpenAQ}}.} \bibinfo{year}{2023}\natexlab{}.
\newblock
\newblock
\urldef\tempurl%
\url{https://openaq.org/}
\showURL{%
\tempurl}


\bibitem[\protect\citeauthoryear{{PurpleAir | Real-time Air Quality
  Monitoring}}{{PurpleAir | Real-time Air Quality Monitoring}}{2022}]%
        {PurpleAir2023}
\bibfield{author}{\bibinfo{person}{{PurpleAir | Real-time Air Quality
  Monitoring}}.} \bibinfo{year}{2022}\natexlab{}.
\newblock
\newblock
\urldef\tempurl%
\url{https://www2.purpleair.com/}
\showURL{%
\tempurl}


\bibitem[\protect\citeauthoryear{Vito}{Vito}{2016}]%
        {misc_air_quality_360}
\bibfield{author}{\bibinfo{person}{Saverio Vito}.}
  \bibinfo{year}{2016}\natexlab{}.
\newblock \bibinfo{title}{{Air Quality}}.
\newblock \bibinfo{howpublished}{UCI Machine Learning Repository}.
\newblock
\newblock
\shownote{{DOI}: https://doi.org/10.24432/C59K5F.}


\bibitem[\protect\citeauthoryear{Wu, Pan, Long, Jiang, and Zhang}{Wu
  et~al\mbox{.}}{2019}]%
        {wu2019graph}
\bibfield{author}{\bibinfo{person}{Zonghan Wu}, \bibinfo{person}{Shirui Pan},
  \bibinfo{person}{Guodong Long}, \bibinfo{person}{Jing Jiang}, {and}
  \bibinfo{person}{Chengqi Zhang}.} \bibinfo{year}{2019}\natexlab{}.
\newblock \showarticletitle{Graph wavenet for deep spatial-temporal graph
  modeling}. In \bibinfo{booktitle}{\emph{IJCAI}}. \bibinfo{pages}{1907--1913}.
\newblock


\bibitem[\protect\citeauthoryear{Yoon, Jarrett, and Van~der Schaar}{Yoon
  et~al\mbox{.}}{2019}]%
        {yoon2019time}
\bibfield{author}{\bibinfo{person}{Jinsung Yoon}, \bibinfo{person}{Daniel
  Jarrett}, {and} \bibinfo{person}{Mihaela Van~der Schaar}.}
  \bibinfo{year}{2019}\natexlab{}.
\newblock \showarticletitle{Time-series generative adversarial networks}.
\newblock \bibinfo{journal}{\emph{NeurIPS}}  \bibinfo{volume}{32}
  (\bibinfo{year}{2019}).
\newblock


\bibitem[\protect\citeauthoryear{Zheng, Liu, and Hsieh}{Zheng
  et~al\mbox{.}}{2013}]%
        {zheng2013u}
\bibfield{author}{\bibinfo{person}{Yu Zheng}, \bibinfo{person}{Furui Liu},
  {and} \bibinfo{person}{Hsun-Ping Hsieh}.} \bibinfo{year}{2013}\natexlab{}.
\newblock \showarticletitle{U-air: When urban air quality inference meets big
  data}. In \bibinfo{booktitle}{\emph{KDD}}. \bibinfo{pages}{1436--1444}.
\newblock


\bibitem[\protect\citeauthoryear{Zuo, Zeitouni, and Taher}{Zuo
  et~al\mbox{.}}{2021}]%
        {zuo2021smate}
\bibfield{author}{\bibinfo{person}{Jingwei Zuo}, \bibinfo{person}{Karine
  Zeitouni}, {and} \bibinfo{person}{Yehia Taher}.}
  \bibinfo{year}{2021}\natexlab{}.
\newblock \showarticletitle{Smate: Semi-supervised spatio-temporal
  representation learning on multivariate time series}. In
  \bibinfo{booktitle}{\emph{2021 IEEE International Conference on Data Mining
  (ICDM)}}. IEEE, \bibinfo{pages}{1565--1570}.
\newblock


\bibitem[\protect\citeauthoryear{Zuo, Zeitouni, Taher, and
  Garcia-Rodriguez}{Zuo et~al\mbox{.}}{2023}]%
        {zuo2023graph}
\bibfield{author}{\bibinfo{person}{Jingwei Zuo}, \bibinfo{person}{Karine
  Zeitouni}, \bibinfo{person}{Yehia Taher}, {and} \bibinfo{person}{Sandra
  Garcia-Rodriguez}.} \bibinfo{year}{2023}\natexlab{}.
\newblock \showarticletitle{Graph convolutional networks for traffic
  forecasting with missing values}.
\newblock \bibinfo{journal}{\emph{Data Mining and Knowledge Discovery}}
  \bibinfo{volume}{37}, \bibinfo{number}{2} (\bibinfo{year}{2023}),
  \bibinfo{pages}{913--947}.
\newblock


\end{thebibliography}

\end{document}